\documentclass[useAMS]{statsoc}

\usepackage{graphicx}
\usepackage{amsmath,amssymb}
\usepackage{mathtools}
\usepackage{bbm}
\usepackage{booktabs}
\usepackage[figuresleft]{rotating}
\usepackage{multirow}
\setcitestyle{aysep={}} % no comma between author and year
\usepackage[caption = false]{subfig} 
\usepackage{color}
\usepackage{tabularx}
\usepackage{setspace}
\usepackage{hyperref}

\usepackage[a4paper]{geometry}
\usepackage{natbib}

\usepackage{etoolbox}
\makeatletter
\patchcmd{\@makecaption}
  {\parbox}
  {\advance\@tempdima-\fontdimen2} % decrease the width!
  {}{}
\makeatother

\title[Time-dependent Cox Survival Neural Network]{tdCoxSNN: Time-Dependent Cox Survival Neural Network for Continuous-time Dynamic Prediction}
\author[Zeng et al.]{Lang Zeng$^{1}$, 
Jipeng Zhang$^{1}$,
Wei Chen$^{2}$, and Ying Ding$^{1,*}$}
\address{$^{1}$Department of Biostatistics, University of Pittsburgh, Pittsburgh, PA, U.S.A.\\
$^{2}$Department of Pediatrics, University of Pittsburgh, Pittsburgh, PA, U.S.A.\\
*Corresponding Author Email: yingding@pitt.edu}
\begin{document}

\begin{abstract}
The aim of dynamic prediction is to provide individualized risk predictions over time, which are updated as new data become available. In pursuit of constructing a dynamic prediction model for a progressive eye disorder, age-related macular degeneration (AMD), we propose a time-dependent Cox survival neural network (tdCoxSNN) to predict its progression using longitudinal fundus images. tdCoxSNN builds upon the time-dependent Cox model by utilizing a neural network to capture the non-linear effect of time-dependent covariates on the survival outcome. Moreover, by concurrently integrating a convolutional neural network (CNN) with the survival network, tdCoxSNN can directly take longitudinal images as input. We evaluate and compare our proposed method with joint modeling and landmarking approaches through extensive simulations. We applied the proposed approach to two real datasets. One is a large AMD study, the Age-Related Eye Disease Study (AREDS), in which more than 50,000 fundus images were captured over a period of 12 years for more than 4,000 participants. Another is a public dataset of the primary biliary cirrhosis (PBC) disease, where multiple lab tests were longitudinally collected to predict the time-to-liver transplant. Our approach demonstrates commendable predictive performance in both simulation studies and the analysis of the two real datasets. %tdCoxSNN has been implemented in PyTorch, Tensorflow, and R-Tensorflow. % The code is available at \href{https://github.com/langzeng/tdCoxSNN}{https://github.com/langzeng/tdCoxSNN}.
\\

\end{abstract}


\maketitle

\section{Introduction}
\label{s:intro}
For many chronic progressive diseases, the prognosis and severity of the disease change over time. A dynamic prediction model that can update the predicted longitudinal disease progression profile with newly acquired data is a critical and unmet need \citep{jenkins2018dynamic}. The unstructured observation
times and the varying number of observations among subjects make building a dynamic prediction model challenging. The complexity is further heightened by the collection of high-dimensional longitudinal data, necessitating the development of innovative dynamic prediction models that can handle large inputs, such as images.

Joint modeling (JM) is a popular approach for dynamic prediction. It consists of a sub-model for the longitudinal process and a sub-model for the survival process linked through a shared pattern  \citep{tsiatis2004joint,rizopoulos2011dynamic,rizopoulos2014r,li2019dynamic,mauff2020joint}. In recent years, functional joint models have been proposed to model the underlying functional form of longitudinal covariates over time \citep{li2022joint,zou2023multivariate}. For example, \citet{zou2023multivariate} applied the functional JM model to predict the progression of Alzheimer’s disease using pre-specified voxels from longitudinal MRI scans. However, this methodology heavily depends on image registration and pre-processing, and it tends to overlook the correlation between voxels. Meanwhile, JM is computationally expensive and especially challenging to directly model high-dimensional longitudinal features such as images.  

%The Bayesian approach is a popular method for parameter estimation in JM \citep{rizopoulos2011dynamic}. After estimating the parameters, the prediction of the survival probability $\hat{\pi}_j(u|s)$ is obtained from the posterior distribution of the predicted survival process \citep{rizopoulos2014r}. 

% In contrast, landmarking avoids directly modeling the longitudinal covariates. Instead, it estimates the effect of predictors through a survival model encompassing all at-risk subjects at a specified landmark time point. Only the observations predating the landmark time are utilized for model training \citep{van2007dynamic}. 

Another popular dynamic prediction approach is the landmarking method (LM), which does not model the longitudinal process. The strategy is to directly model the bilateral relationship between predictors at the landmark time and the corresponding residual survival time (i.e., the observed survival time subtracts the landmark time) \citep{van2007dynamic}. To obtain the conditional survival probability beyond the landmark time, the model is fitted only using subjects who are at risk at the landmark time. Hence, the estimated effects of predictors at landmark time can approximate the effect of the time-varying covariates on survival outcomes beyond the landmark time. To avoid loss of longitudinal
information before the landmark time, \citet{paige2018landmark} proposed to separately model the longitudinal measurements first, and then use the predicted of value of covariates (instead of observed) at the landmark time for the survival model. However, selecting the appropriate landmark time can be a challenging task. While the concept of a landmarking supermodel, as introduced by \citet{van2011dynamic}, offers a solution to select the best landmark time from multiple candidate landmark times, it may not fully address the complexities of real-world scenarios. In practical applications, a more flexible model that can accommodate unpredictable visiting times, the points at which predictions are updated, is often preferable.

With the recent advancement of machine learning and its success in analyzing survival data \citep{katzman2018deepsurv,lee2018deephit,kvamme2019time,lin2021functional,zhong2022deep}, new methods have been developed to integrate the dynamic prediction models with machine learning techniques to expand their applications and enhance the prediction accuracy in more complex settings. For example, \citet{tanner2021dynamic} combined LM with machine learning ensembles to integrate prediction from multiple methods to achieve improved performance. Moreover, \citet{lee2019dynamic,jarrett2019dynamic,nagpal2021deep,lin2022deep} proposed different deep learning models for dynamic predictions on the discrete-time scale. Among them, Dynamic-DeepHit \citep{lee2019dynamic} the most popular method for dynamic prediction, serving as an extension of the original DeepHit model that directly models the conditional probabilities to survive beyond each of the pre-specified discrete time points. Although discretizing the time does not necessarily reduce prediction accuracy, the number of time intervals used for discretization significantly impacts accuracy, so it needs to be carefully tuned in practice \citep{sloma2021empirical}. In summary, there is a lack of dynamic prediction models that could directly handle the high-dimensional, longitudinal variables collected at unstructured observational times on a continuous scale.

In this paper, we propose a dynamic prediction method by integrating the time-dependent Cox model \citep{fisher1999time,thomas2014tutorial} with the neural network. This continuous-time method can characterize the non-linear relationship between the longitudinal and time-to-event processes. It fully uses the available longitudinal predictors to train the prediction model without explicitly modeling the longitudinal process. The proposed method can also incorporate large-dimensional inputs (e.g., images) through fitting additional neural network architectures (e.g., convolution neural network or recurrent neural network) simultaneously with the survival neural network.

\section{Dynamic prediction using time-dependent Cox survival neural network}\label{proposedmodel}
\subsection{Notation}
Let $D_n=\{T_i,\delta_i,X_i,\mathcal{Y}_i(T_i); i = 1,\dots,n\}$ denote the data from $n$ samples. $T_i = \min(T_i^*,C_i^*)$ denotes the observed event time for subject $i$, where $T_i^*$ and $C_i^*$ are the underlying true event time and censoring time, and $\delta_i = I(T_i^* \leq C_i^*)$ is the event indicator. $X_i$ is the time-invarying covariate for subject $i$. We let $y(t)$ denote the longitudinal measurement taken at time point $t$. $\mathcal{Y}_i(t) = \{ y_i(t_{il}) | 0\le t_{il} \le t , l = 1,\dots ,n_{i}\}$ are the $n_i$ longitudinal measurements for subject $i$ between time 0 and $t$. %We allow the number of longitudinal observations ($n_i$) to differ across individuals.

For dynamic prediction, we are interested in predicting the probability that a new patient $j$, with time-varying measurements up to time $s$, will survive beyond time $u$ for $u>s$, denoted as 
$\pi_j(u|s)=Pr(T_j^*> u|T_j^*>s,\mathcal{Y}_j(s),X_j).$
In contrast to static predictions, dynamic models allow predictions to
be updated and obtain $\hat{\pi}_j(u|s')$  when new information is collected at time $s'>s$. 

\subsection{Model and Estimation}
The time-dependent Cox model assumes
\begin{equation}
h_i(t|X_i,\mathcal{Y}_i(t)) =  h_0(t)\exp[\beta^TX_i+\gamma ^Ty_i(t)], \label{eq:tdCox}
\end{equation}
where $h_0(t)$ is the baseline hazard function and $h_i(t|X_i,\mathcal{Y}_i(t))$ is the hazard function of the $i$th individual with covariates $X_i$ and $\mathcal{Y}_i(t)=\{y_i(t^\prime), 0\le t^\prime \le t\}$. The coefficients $(\beta,\gamma)$ are estimated through the partial likelihood
$$\prod_i^n \left\{ \frac{\exp[g_{\beta,\gamma}(X_i,y_i(T_i))]}{\displaystyle\sum_{j:T_j\geq T_i} \{ \exp[g_{\beta,\gamma}(X_j,y_j(T_i))]-E_j(g_{\beta,\gamma}) \} } \right\} ^{\delta_i},$$ 
where $g_{\beta,\gamma}(X_i,y_i(T_i)) = \beta^TX_i+\gamma ^Ty_i(T_i)$ is the risk score function and
$$E_j(g) = \frac{\sum_{k}\mathbbm{1}(k>j, T_k= T_j)}{\sum_{k}\mathbbm{1}(T_k= T_j)}\sum_{k: T_k=T_j} \exp[g(X_k,y_k(T_j))]$$ is Efron's approximation for handling tied events \citep{efron1977efficiency}. Note that, in general, the time-dependent predictors used in this model (\ref{eq:tdCox}) could be a summary measure of the history (from time 0 to $t$). For example, one may use the mean or maximum of the longitudinal process $\mathcal{Y}_i(t)=\{y_i(t^\prime), 0\le t^\prime \le t\}$ as a summary measure. The choice depends on the research problem and is well discussed in \cite{fisher1999time}. In this work, we use the instantaneous measurement at time $t$ (i.e., $y_i(t)$) as the time-dependent predictor in the model (\ref{eq:tdCox}). % to handle the nonlinear effect of covariates and incorporate the high-dimensional longitudinal data.

We expand the time-dependent Cox model using a neural network to replace the linear effect of the covariates in the model (\ref{eq:tdCox}). Specifically, the time-dependent Cox survival neural network (tdCoxSNN) is a feed-forward neural network that models the nonlinear effect of time-dependent covariates on the hazard function. Instead of assuming the risk score function $g(X_i,y_i(t))$ takes the linear form, we leave the form of $g(X_i,y_i(t))$ unspecified and model it through a neural network $g_\theta(X_i,y_i(t))$, parameterized by $\theta = (\mathbf{W},\mathbf{V})$, where $\mathbf{W}$ is the weight matrices and $\mathbf{V}$ is the shift vectors of the neural network. The input of the neural network is the predictors $(X_i,y_i(t))$ and the output is a single node that represents the risk score $g_\theta(X_i,y_i(t))$. The parameter $\theta$ is estimated by minimizing the negative log partial likelihood $\hat\theta = \arg\min_{\theta} l(\theta),$ where  
\begin{equation}
\label{eq:lpl}
 l(\theta) =  -\frac{1}{n}\sum_{i=1}^n\delta_i \left[g_\theta(X_i,y_i(T_i))-\log\left\{\sum_{j:T_j\geq T_i} \exp(g_\theta(X_j,y_j(T_i)))-E_i(g_\theta)\right\} \right].
\end{equation}

Note that at each event time $T_i$, $y_j(T_i)$ is required for all the subjects who are at risk at $T_i$ (i.e., $j: T_j \geq T_i$) in (\ref{eq:lpl}), which is not always available in practice. Therefore, interpolation between longitudinal measurements is required. In the standard time-dependent Cox model, a commonly used strategy is to assume the values of the longitudinal predictors stay constant until the next measurement becomes available. Specifically, in the calculation of the loss function (\ref{eq:lpl}), we use $y_j(T_i^-)$, which are the longitudinal measures immediately before $T_i$. This is consistent with how the standard time-dependent Cox model is implemented in R and SAS \citep{thomas2014tutorial,therneau2017using}.

Besides the ability to model complex nonlinear effects, another advantage of using the neural network is that a pre-trained neural network, which models a specific data type such as image or text, can be readily combined with tdCoxSNN. For example, ResNet50 \citep{he2016deep} is a convolutional neural network (CNN) for image classification whose weights have been trained over one million images. Adding the pre-trained CNN on top of tdCoxSNN allows the method to take the images directly as part of the input, which makes the large image data processing and the survival data modeling (with image data as predictors) an end-to-end procedure. 

% subject time1 time2 death creatinine 
%      15     0    90     0        0.9
% The underlying code treats intervals as open on the left and closed on the right,e.g. the creatinine on exactly day 90 is 0.9.

\begin{figure}
\centering
\subfloat[\label{fig:Ng1}]{%
    \includegraphics[width=1.1\linewidth]{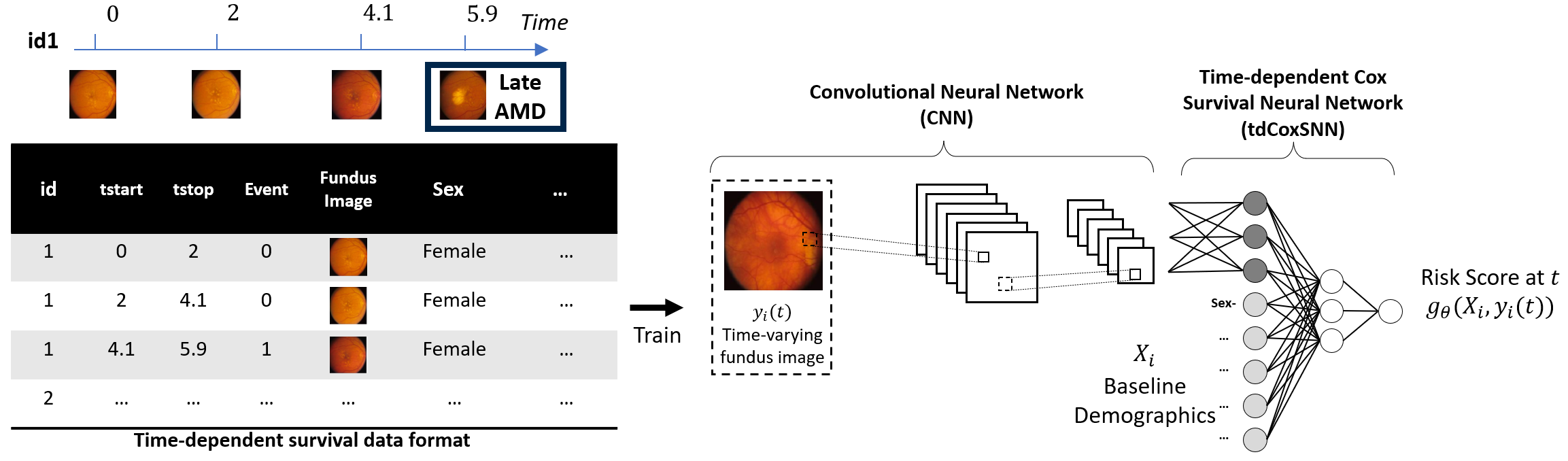}
}    
\hfill
\subfloat[\label{fig:Ng2}]{%
    \includegraphics[width=0.99\linewidth]{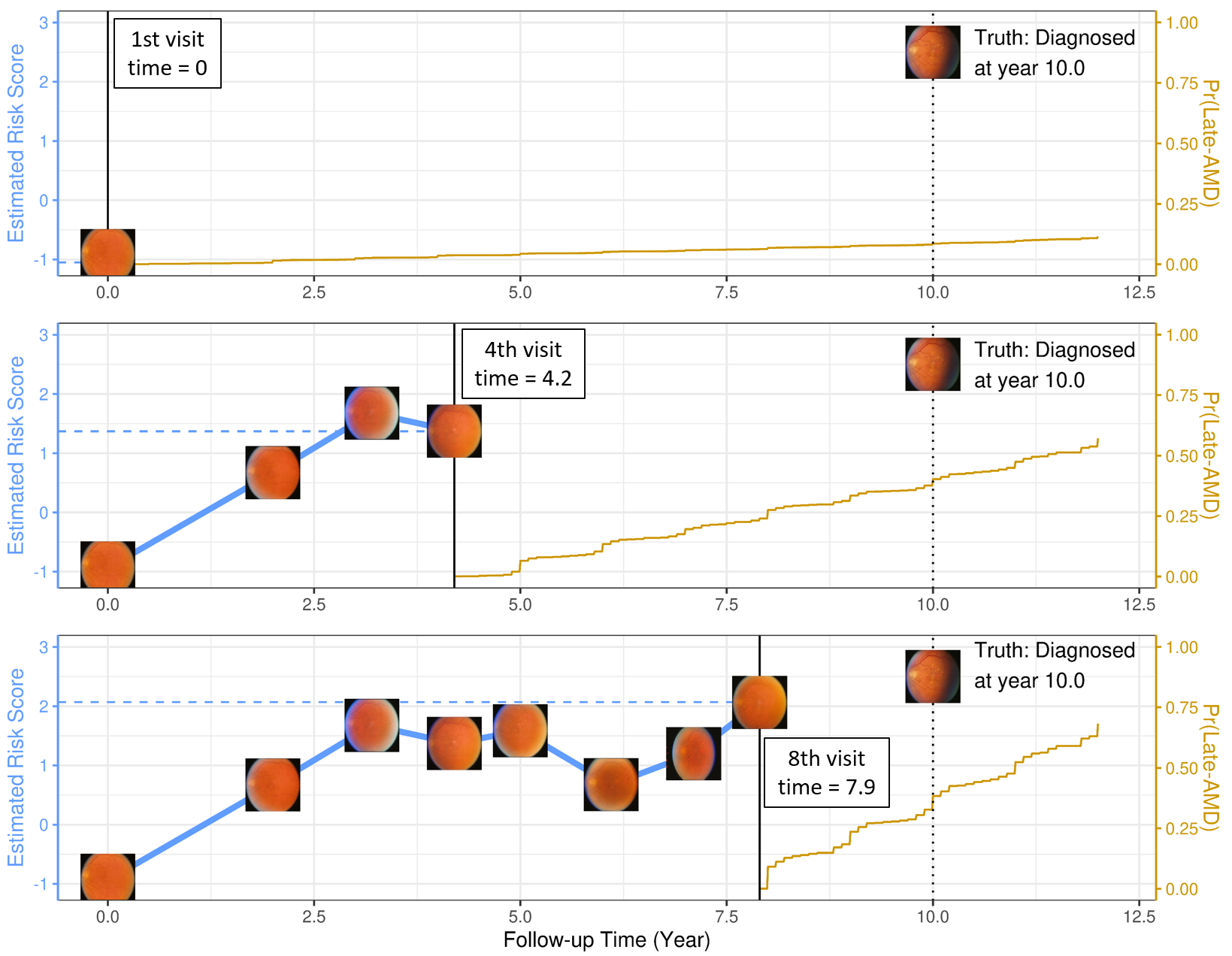}
}
\caption[Two numerical solutions]{\label{workflow}\textbf{(a)} Workflow of tdCoxSNN illustrated using the AMD example. \textbf{(b)} Dynamic prediction using tdCoxSNN for the AMD example. The y-axis on the left denotes the risk score estimated from the tdCoxSNN at each visit. The y-axis on the right is the predicted probability of developing late AMD since the latest visit time.}
\end{figure}

Here, we use our first application to demonstrate the workflow of tdCoxSNN in Figure \ref{workflow}a. The goal is to establish a dynamic prediction model for the progressive eye disease, age-related macular degeneration (AMD), using longitudinal fundus images, where the outcome of interest is time-to-late-AMD. We first convert the longitudinal observations into a long data format, indexed by intervals (tstart, tstop] that separate the time periods between longitudinal measurements for each individual. Within each interval, the longitudinal variable (i.e., fundus image) represents the measurement recorded at ``tstart", and the event indicator is determined based on the event status at ``tstop". The entire neural network takes fundus images and other predictors as inputs, where images are handled by a CNN with output nodes fed into tdCoxSNN together with other predictor nodes to estimate the risk scores. The loss function (\ref{eq:lpl}) is computed based on these risk scores, and the parameters of the neural network are optimized accordingly. The risk score associated with each time interval (tstart, tstop] assumes that the value remains constant throughout the interval.

\subsection{Prediction using tdCoxSNN\label{tdCoxSNN_predict}}
Once we trained the tdCoxSNN model, the predicted probability for a new coming subject $j$ with information up to time $s$ is given by 
\begin{equation}\label{eq:p}
\hat\pi_j(u|s)=\exp\left [-\{\hat H_0(u;g_{\hat{\theta}})-\hat H_0(s;g_{\hat{\theta}})\}\exp\{g_{\hat{\theta}}(X_j,y_j(s))\}\right ] \text { for } u>s,
\end{equation}
where $\hat H_0(t;g_{\hat{\theta}})$ is the Breslow estimator of the cumulative baseline hazard function with $g_{\hat{\theta}}$ from the fitted neural network in Figure \ref{workflow}a. $\hat\pi_j$ can be updated when new information of subject $j$ is available at a later time point $s^\prime > s$. 
Figure \ref{workflow}b) uses the AMD application to illustrate the dynamic prediction. Initially, this new patient only had a fundus image taken at time 0, which is used to predict the probability of a late AMD progression curve since time 0. As this patient underwent subsequent visits with additional fundus images taken (shown in the second and third panels in Figure \ref{workflow}b, the predicted probability curve since the latest visit time is updated using the latest fundus image.

Note that the estimator of $\theta$, which can minimize the loss (\ref{eq:lpl}), may not be unique. It is possible to find $\Tilde{\theta}$ such that $g_{\Tilde{\theta}} := g_{\hat\theta}+c$ with $c$ being a constant. Note that $\Tilde{\theta}$ is also a minimizer since $l(\Tilde{\theta}) = l(\hat\theta)$. However, the constant shift $c$ of the estimated hazard function will not change the predicted probability in (\ref{eq:p}) as it appears both in the numerator and the denominator. Therefore, $\hat\pi_j(u|s)$ in (\ref{eq:p}) is robust to a constant shift of the neural network output. 

A key observation is that fitting a time-dependent Cox model is essentially equivalent to fitting a time-independent Cox model with delayed time entry (aka, left truncation). Consider the subject id = 1 in Figure \ref{workflow}a. This individual has three longitudinal measurements recorded at times t = 0, 2, 4.1, and is diagnosed with late-AMD at $t=4.9$. In the context of the loss function (\ref{eq:lpl}), these three rows are treated as three separate ``pseudo'' subjects. Within the cohort, these pseudo subjects are at risk during the intervals [0, 2), [2, 4.1), and [4.1, 5.9), with the last two having delayed entry at times 2 and 4.1, respectively. Therefore, the time-dependent Cox model is essentially a standard time-independent Cox model that treats multiple intervals as multiple pseudo subjects with delayed time entry. Our method can be viewed as a time-independent model that evaluates the impact of predictors at the current time on future survival processes. 

Here, we elaborate on the connection and distinction between this time-dependent Cox model-based prediction and the standard LM-based prediction. Note that the prediction equation (\ref{eq:p}) aligns with the prediction formula used in LM (see equation (5) in \citet{tanner2021dynamic}) under the assumption that our baseline hazard at any future time ($u>s$) and the effect of predictors do not change for different choices of landmark times. In terms of distinction, LM requires to re-train the model using the subset of samples who survived beyond the landmark time. It also only uses the single measurement recorded immediately before or at the landmark time while ignoring the rest of the longitudinal measures. Instead, our method uses all training samples with all their longitudinal measurements up to their event or censoring time.

% The assumption of time-invariant covariate effects can be relaxed by incorporating time as an additional input into the neural network, (e.g., $g_\theta(X_i,y_i(s),t)$). This modification allows the model to account for changes in predictor effects over time, enhancing its ability to adapt to dynamic patterns in the data \citep{kvamme2019time}.

%In the connection either between the standard Cox model and landmarking, the time-dependent Cox model can be viewed as a time-independent model that evaluates the impact of current predictors on future disease progression. Therefore, we anticipate that equation (\ref{eq:p}) will provide reasonable predictions. This expectation is confirmed by the robust and satisfactory prediction outcomes observed in various simulation scenarios and real data applications, demonstrating the model's power in predicting disease progression, especially with longitudinal high-dimensional predictors.

\section{Prospective accuracy metrics}\label{metric}
Methods for assessing the predictive performance of survival models concentrates either on calibration, i.e., how well the model predicts the observed data, or on discrimination, i.e., how well the model discriminates between subjects with the event from subjects without. We consider both calibration and discrimination metrics for model performance evaluation. Similar to previous studies for joint modeling and landmarking, we compare dynamic prediction models on a pre-specified time-window $(s,s+\Delta t]$ where $s$ is the landmark time \citep{rizopoulos2011dynamic,rizopoulos2017dynamic,tanner2021dynamic}.

\subsection{Calibration metric: time-dependent Brier Score\label{sec:bs}}
The time-dependent Brier Score measures the mean squared error between the observed survival status and the predicted survival probability weighted by the inverse probability of censoring (IPCW) \citep{gerds2006consistent}. A lower Brier Score indicates a higher prediction accuracy. For a given landmark time $s$, the estimated Brier Score at time $s+\Delta t$ is 
$$
\widehat{BS}(s,\Delta t;\hat\pi) = \frac{1}{\sum_{i}\mathbbm{1}(T_i>s)}\sum_{i=1}^n\left[  \mathbbm{1}(T_i>s)\hat W_i(s,\Delta t) \{\mathbbm{1}(T_i>s+\Delta t)-\hat \pi_i(s+\Delta t|s)\}^2 \right],
$$
where $ \hat W_i(s,\Delta t) = \frac{\mathbbm{1}(T_i>s+\Delta t)}{\hat G(s+\Delta t|s)}+\frac{\mathbbm{1}(T_i\leq s+\Delta t)\delta_i}{\hat G(T_i^-|s)}$ is the IPCW weight and $\hat G(u|s) = \frac{\hat G(u)}{\hat G(s)}$ is the Kaplan-Meier estimate of the conditional censoring distribution $Pr(C^*>u|C^*>s)$ with $u>s$.

\subsection{Discrimination metric: time-dependent AUC}
The area under the receiver operating characteristic curve (AUC) measures the discrimination of the prediction model. It ranges from 0 to 1, and AUC close to 1 suggests better discrimination. We use cumulative sensitivity and dynamic specificity AUC (cdAUC) \citep{kamarudin2017time} to evaluate the discrimination performance of the models at different time points. Given a predictor $X$ and a threshold $b$ and assuming the larger value of $X$ is associated with shorter survival time, the cumulative sensitivity is defined as $Se^C(b,\Delta t)=P(X_i>b|s<T_i\leq s+\Delta t)$ while the dynamic specificity is $Sp^D(b,\Delta t)=P(X_i\leq b|T_i>s+\Delta t)$. The term ``cumulative" is used to differentiate this sensitivity from the incident sensitivity $Se^I(b,\Delta t)=P(X_i>b|T_i=s+\Delta t)$ which assesses the sensitivity for the population whose survival time exactly equals $s+\Delta t$. With the cumulative sensitivity and dynamic specificity, the corresponding $\text{cdAUC}(s,\Delta t)$ is defined as $ \text{cdAUC}(s,\Delta t) = P(X_i > X_j|s < T_i\leq s+ \Delta t, T_j>s+\Delta t)$, $i \neq j$. Specifically, for a given time interval $(s,s+\Delta t]$, the IPCW estimator of cdAUC is $$\widehat{\text{cdAUC}}(s,\Delta t) = \frac{\displaystyle\sum_{i=1}^n\displaystyle\sum_{j=1}^n \mathbbm{1}_{(\hat \pi_i(s+\Delta t|s)<\hat \pi_j(s+\Delta t|s))}\delta_i\mathbbm{1}_{(s<T_i\leq s+\Delta t)}\mathbbm{1}_{(T_j> s+\Delta t)} W_i(s,\Delta t)W_j(s,\Delta t)}{\displaystyle\sum_{i=1}^n\displaystyle\sum_{j=1}^n \delta_i\mathbbm{1}_{(s<T_i\leq s+\Delta t)}\mathbbm{1}_{(T_j> s+\Delta t)}W_i(s,\Delta t)W_j(s,\Delta t)},$$
where $W_i(s,\Delta t)$ is defined in Section \ref{sec:bs}. It computes the IPCW weighted percentage of the comparable subject pairs $(i,j)$ where the order of their predicted survival probabilities is consistent with their observed time for the given time interval $(s, s+\Delta t]$. The comparable pair $(i,j)$ contains two subjects in which subject $i$ experiences the event within $(s, s + \Delta t]$ and subject $j$ is event-free by $s+\Delta t$.

% Similarly, the $\Delta t$ IPCW estimator is
% $$\hat{AUC}(\Delta t) = \frac{\sum_i\sum_j \mathbbm{1}(\hat \pi_i(t_i+\Delta t|t_i)<\hat \pi_j(t_j+\Delta t|t_j))D_i(t_i,\Delta t)(1-D_j(t_j,\Delta t))W_i(\Delta t)W_j(\Delta t)}{\sum_i^n\sum_j^n D_i(t_i,\Delta t)(1-D_j(t_j,\Delta t))W_i(\Delta t)W_j(\Delta t)}.$$

\section{Numerical implementation and simulations}\label{simulation}
\label{computationdetails}
\subsection{Implementation}\label{implementation}
For the tdCoxSNN, we implement the log partial likelihood function with Efron's tie approximation (\ref{eq:lpl}) in Tensorflow \citep{abadi2016tensorflow}, PyTorch \citep{paszke2019pytorch}, and R-Tensorflow \citep{allaire2019tensorflow}, which are available at \url{https://github.com/langzeng/tdCoxSNN}. It is implemented through matrix operations, so the calculation is fast (see supplementary materials for details). 
We use the following survival neural network structure in all simulations and real data analysis: input layer $\to$ one hidden layer $\to$ batch normalization layer $\to$ dropout layer $\to$ output layer. The batch normalization layer \citep{ioffe2015batch} accelerates the neural network training, and the dropout layer \citep{srivastava2014dropout} protects the neural network from over-fitting. Hyper-parameters are also fixed in all analyses: 30 nodes in the hidden layer, Scaled Exponential Linear Unit (SeLU) as the hidden layer activation function, batch size 50, epoch size 20, learning rate 0.01, and dropout rate 0.2. The Adam optimizer is used to optimize the neural network \citep{kingma2014adam}. 

\subsection{Simulation 1: low-dimensional predictors}\label{simulation1}
We carried out extensive simulations to empirically assess and compare the performance of the proposed tdCoxSNN with other four dynamic prediction models. In the first low-dimensional setting, we simulated the data with four time-invarying covariates and one longitudinal covariate. Survival and longitudinal data were generated through the joint models \citep{arisido2019joint} with the longitudinal covariate $y(t)$ is either linear or non-linear in time $t$ and the risk score function $g(x,y(t))$ is either a linear or non-linear function of $x$ and $y(t)$. Specifically, for sample $i$, the observed one-dimensional longitudinal covariate $y_i(t)$ was generated through $y_i(t) = y^*_i(t)+\epsilon$ where $y^*_i(t)$ is the true longitudinal trajectory over time and $\epsilon\sim N(0,0.3^2)$ represents a measurement error. The true value of the time-varying covariate is given by
$$y^*_i(t) = \beta_0+\beta_1t+\beta_2t^2+b_{i0}+b_{i1}t+b_{i2}t^2 \text{ and } \mathbf{b}_i = (b_{i0},b_{i1},b_{i2})^T\sim N(0,\Sigma_{3\times 3})$$
where $\beta_0 = 3.2$, $\beta_1=-0.07$. $\Sigma$ denotes a 3 by 3 inter-subject variance matrix with $\Sigma_{11} = 1.44, \Sigma_{22}=0.6$ and we assume the covariances $\Sigma_{ij}$ are zero in all simulations. $(\beta_2, \Sigma_{33})$ captures the non-linearity of the trajectory, which is $(\beta_2,  \Sigma_{22}) = (0.004, 0.09)$ for the non-linear case and $=(0,0)$ for the linear case. $y_i(t)$ was measured at each $t=0,1,2,\dots,14$. The survival time $T_i^*$ was obtained through a Weibull model $h_i(t) =\lambda \rho t^{\rho-1}\exp\{g(X_i,y^*_i(t))\}$ with $\rho=1.4$ and $\lambda = 0.1$. Specifically, the survival time was generated by evaluating the inverse of the cumulative density function. Since this does not lead to a closed-form expression, we used the R function \{uniroot\} to generate numerically. The censoring time $C_i^*$ was generated through an exponential distribution $\text{exp}(1/7)$. The observed survival time $T_i = \min(T^*_i,C_i^*)$ and the event indicator $\delta_i = \mathbbm{1}(T^*_i\leq C_i^*)$ were then calculated. Longitudinal values $y_i(t)$ with $t\geq T_i$ were disregarded, and only $y_i(t)$ measured before $T_i$ were kept as the observed longitudinal measurements for subject $i$. The baseline covariates $x_k$ $(k=1,2,3,4)$ were independently generated from the continuous uniform distribution on $[-0.5, 1.5]$. We considered four different settings (see below) with the linear or nonlinear risk function $g(X,y^*(t))$ and the linear or quadratic trend in time of the true longitudinal variable $y^*(t)$. We added the intercept term $c$ in the risk score term to adjust the censoring rate.

\begin{enumerate}
    \item Linear Effect + Linear Trend: \\
  $\begin{cases}
  g(X,y^*(t)) = x_1+2x_2+3x_3+4x_4+0.3y^*(t)+c \\
  y^*(t)=(3.2+b_0)-(0.07+b_1)t   
  \end{cases}$
    \item Linear Effect + Non-linear Trend: \\$\begin{cases}
  g(X,y^*(t)) = x_1+2x_2+3x_3+4x_4+0.3y^*(t)+c \\
  y^*(t)=(3.2+b_0)-(0.07+b_1)t+(0.004+b_2)t^2
 \end{cases}$
    \item Non-linear Effect + Linear Trend: \\$\begin{cases}
  g(X,y^*(t)) = \frac{1}{3} [x_1^2x_2^3+\log(x_3+1)+\{0.3y^*(t)x_4+1\}^{\frac{1}{3}}+\exp(\frac{x_4}{2})+0.3y^*(t)]^2+c \\
  y^*(t)=(3.2+b_0)-(0.07+b_1)t 
 \end{cases}$
    \item Non-linear Effect + Non-linear Trend: \\$\begin{cases}
  g(X,y^*(t)) = \frac{1}{3}[x_1^2x_2^3+\log(x_3+1)+\{0.3y^*(t)x_4+1\}^{\frac{1}{3}}+\exp(\frac{x_4}{2})+0.3y^*(t)]^2+c \\
  y^*(t)=(3.2+b_0)-(0.07+b_1)t+(0.004+b_2)t^2 
 \end{cases}$
\end{enumerate}

We implemented five different methods, including the proposed time-dependent Cox survival neural network (tdCoxSNN), the time-dependent Cox model (tdCoxPH), the joint modeling with the time-dependent Cox sub-model for the survival data (JM), the landmarking method with random survival forest (LM-RSF),  and the discrete-time deep learning method (Dyn-DeepHit). For the longitudinal sub-model in JM, we included an intercept and linear time in both fixed effects and random effects parts. We compared these methods across various simulation scenarios, which included different censoring rates (40\% and 80\%) and training sample sizes ($n_{train}$ = 500, 1000). The predictive performance was assessed using an independent test dataset with sample size $n_{test}=200$ at $\Delta t=1,2,3,4$ from the landmark time $s=1$ and 3. For each scenario, we conducted 100 simulation runs.

\begin{figure}
\centering
\includegraphics[width=0.9 \linewidth]{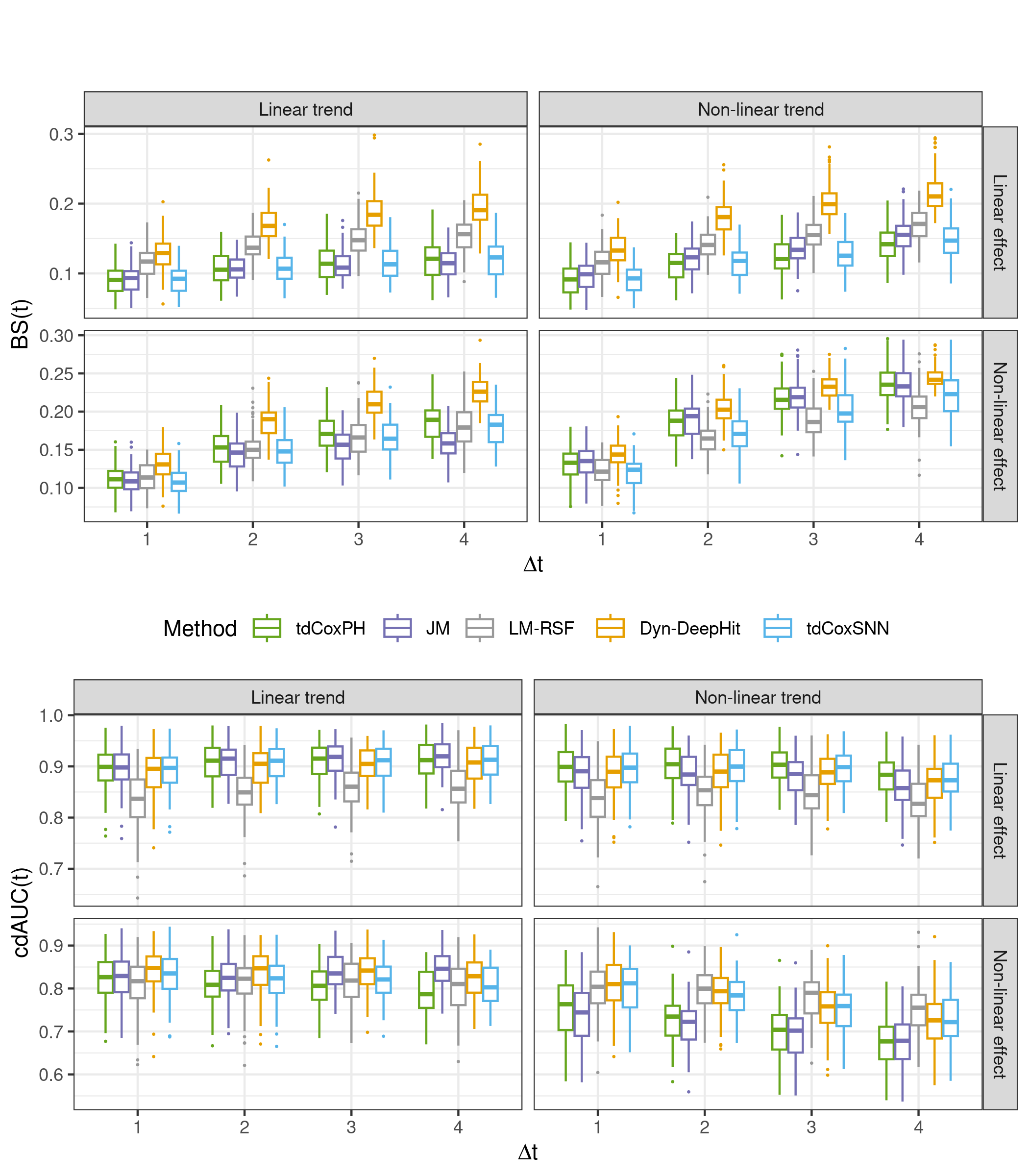}
\caption{Low dimensional simulations: Boxplots of BS and cdAUC at $\Delta t=1,2,3,4$ from landmark time $s=1$ for four simulation settings. Models included in the comparison are the time-dependent Cox model (tdCoxPH), the joint modeling (JM), the LM with RSF (LM-RSF), the Dynamic-DeepHit (Dyn-DeepHit), and the proposed model (tdCoxSNN).}
\label{simulation_low_fig}
\end{figure}

\begin{figure}
\centering
\includegraphics[width=0.9 \linewidth]{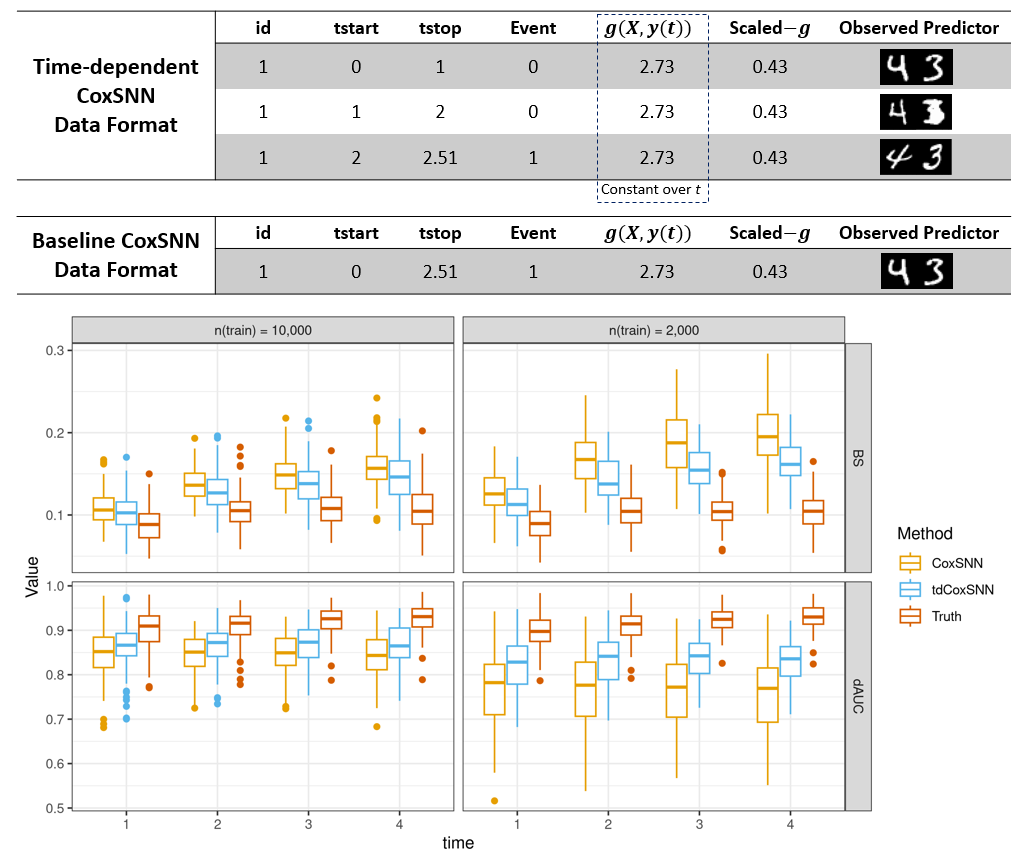}
\caption{High dimensional simulations: Boxplots of BS and cdAUC at $\Delta t=1,2,3,4$ from landmark time $s=1$ in simulation with longitudinal $28\times28$ handwriting digit images. Models included in the comparison are time-independent Cox SNN (CoxSNN), time-dependent Cox SNN (tdCoxSNN), and the true model with the true risk score (Truth).}
\label{simulation_high_fig}
\end{figure}

Figure \ref{simulation_low_fig} presents the boxplots of BS and cdAUC at $\Delta t =$ 1,2,3,4 for all five methods in the scenario with $n_{train} = 500$, 40\% censoring rate, and landmark time $s= 1$. We observed consistent performance patterns in terms of both BS and cdAUC metrics among these five methods. In scenarios where predictor effects are non-linear (lower panel), tdCoxSNN surpasses tdCoxPH in performance. Under linear effect conditions (upper panel), tdCoxSNN shows comparable performance to tdCoxPH. This indicates that the tdCoxSNN is able to learn the nonlinear effect in complex settings while maintaining a similar performance as tdCoxPH when the effect is linear. JM performs the best when both sub-models are correctly specified (left top). However, its prediction accuracy decreases when one or both sub-models are incorrectly specified. This implies the performance of joint modeling highly depends on the correctness of both longitudinal and survival sub-models. %We also noticed that in our simulation, correctly modeling the linear trend of longitudinal can benefit the long-term prediction of JM (left bottom scenario, $\Delta t =4 $) even though the survival sub-model was wrongly specified. 
The LM-RSF performs the worst when the effects of predictors are linear and performs similarly to tdCoxSNN when the effects are nonlinear. The Dyn-DeepHit displays a similar or slightly better prediction accuracy than tdCoxSNN in terms of cdAUC. However, we observed a large calibration error (BS) for Dyn-DeepHit. This is because the loss function in Dyn-DeepHit contains a part that measures the prediction concordance, which is expected to show an advantage under the cdAUC metric. The results of other simulation scenarios can be found in the supplementary materials and the findings are consistent across different scenarios. Our simulation indicates that the tdCoxSNN achieves robust and satisfactory prediction performance under various low-dimensional time-dependent predictor settings. 

\subsection{Simulation 2: high-dimensional predictors}
We further evaluated the performance of tdCoxSNN with high-dimensional predictors. We only considered the Linear Effect + Linear Trend setting as in the low-dimensional setting. Specifically, at each visit time, we mapped the risk score $g(X,y^*(t))$ to a handwritten digit image from the Modified National Institute of Standards and Technology (MNIST) database \citep{lecun1998mnist}. Specifically, we scaled $g(X,y^*(t))$ within $[0,0.99]$ and rounded it to 2 decimal places. Then, the two handwritten digit images (representing the tenth and hundredth numbers) were randomly sampled in the corresponding digit class of MNIST, with around 7,000 corresponding images for each digit. The two $28\times28$ (pixels) images are treated as the observed predictors at time $t$. Simulations were performed in two scenarios. In scenario 1, the true risk score is constant over time, while the mapped longitudinal images vary since the mapping of risk scores to images was performed separately on each visit. In scenario 2, the true risk score and the corresponding longitudinal images both vary over time ($y^*(t)$ is a linear function of $t$). We performed 100 simulation runs for each case with two different sample sizes of training data $n_{train} = 2,000$ and $10,000$. The prediction accuracy metrics $\widehat{\text{BS}} (s,\Delta t)$ and $\widehat{\text{cdAUC}}(s,\Delta t)$ were evaluated on a separate test samples with $n_{test}=200$ at $s = 1$ with $\Delta t=1,2,3,4$. 

The convolutional layers and max pooling layers were added to the survival neural network structure to model the images (Details of the CNN model can be found in the supplementary materials). The tdCoxSNN used all the longitudinal images (up to the observed time $T_i$) to train the model, while the standard time-independent Cox SNN model (CoxSNN) only used the baseline images to train. When making predictions, both tdCoxSNN and CoxSNN used the image at the landmark time. Specifically, tdCoxSNN used the estimated baseline hazard on $(s,s+\Delta t]$, while CoxSNN treated the image taken at the landmark time as the baseline image and used the estimated baseline hazard on $(0,\Delta t]$ to predict future survival probability at $\Delta t$ from the landmark time $s$. The prediction from the true model with the true risk score (Truth) was also presented to represent the best performance one can achieve. Since the other methods cannot directly model the high-dimensional image data, they are not included in this simulation setting.

Figure \ref{simulation_high_fig} presents the results of scenario 1 (i.e., the true risk score is constant over time with 40\% censoring rate) with two different training sample sizes. Since the true risk score is constant over time in this scenario, the essential difference between CoxSNN and tdCoxSNN is that more images were included for training tdCoxSNN. The data input with images were shown in Figure \ref{simulation_high_fig} for both models. We can see that although the risk score $g(X,y^*(t))$ is constant over time, the observed images vary over time due to the random sampling from MNIST. tdCoxSNN outperforms CoxSNN at all $\Delta t$ for both metrics under both sample sizes. When the training sample size is large ($n_{train}=10,000$), tdCoxSNN performs closer to the truth. Scenario 1 demonstrates that tdCoxSNN outperforms CoxSNN even when the risk score is not time-varying. In scenario 2 where the risk scores are time-varying, CoxSNN is less accurate than tdCoxSNN under all settings (see supplementary materials),  which is as expected since CoxSNN ignored longitudinal images and hence failed to capture the effects of time-varying predictors accurately. 

\section{Real data analysis}\label{realdata}

\begin{figure}
\centering
\subfloat[\label{AREDS}]{%
  \includegraphics[width=1.1\linewidth]{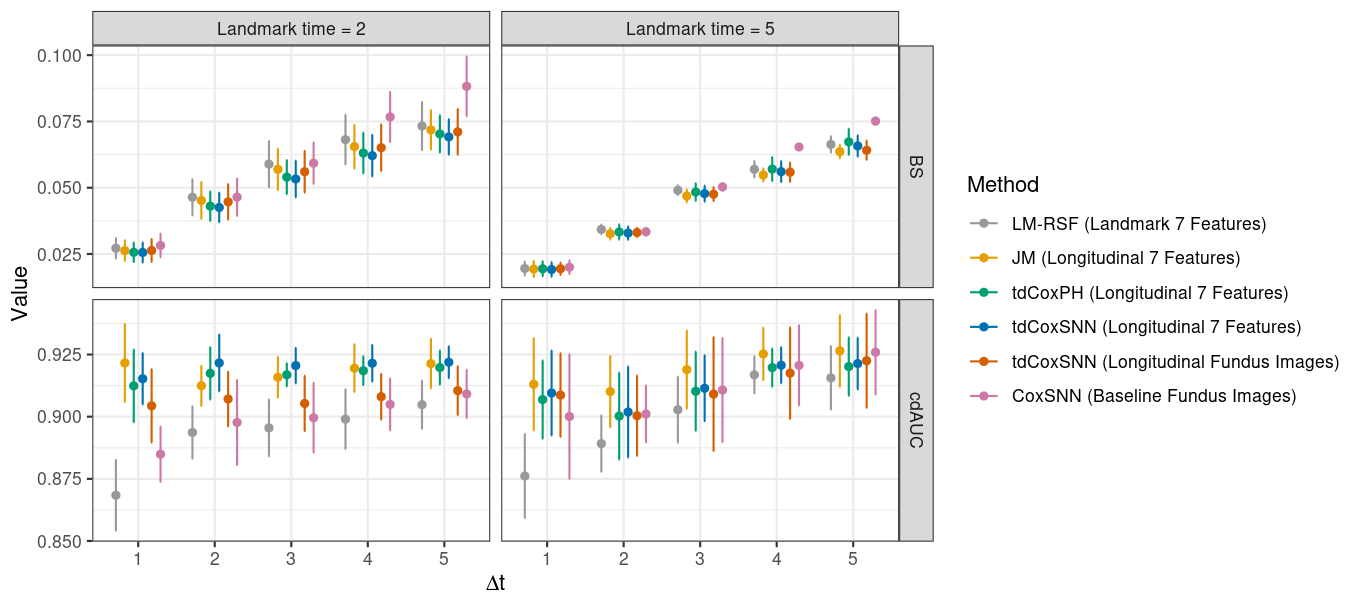}
}
\hfill
\subfloat[\label{saliencymap}]{%
    \includegraphics[width=4in]{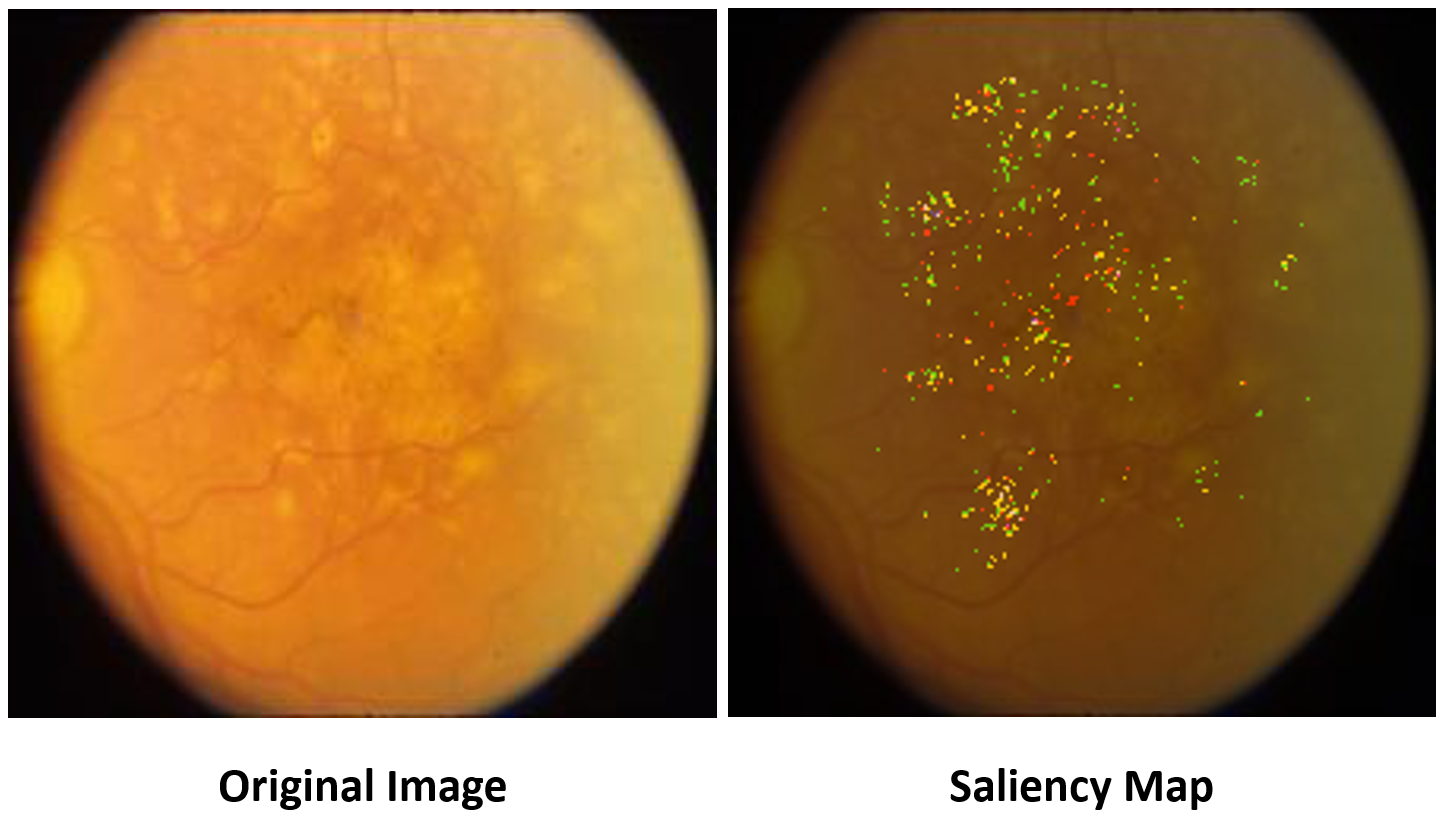}
}

\caption{\label{RealData5foldboxplot_LM}\textbf{(a)} Mean and standard deviation of BS and cdAUC for the 5-fold cross-validation analysis of AREDS data analysis. 
% Static prediction models were fitted on baseline demographics with or without 7 baseline manually extracted image features. 
\textbf{(b)} Saliency map for the left eye at year 2.3 of the subject with baseline age 71.4, at least high-school education, and was a smoker at baseline.}
\end{figure}

\subsection{Application to AMD progression prediction\label{sec:amd}}
We applied the proposed tdCoxSNN on Age-Related Eye Disease Study (AREDS) data and built a dynamic prediction model for time-to-progress to the late stage of AMD (late-AMD) using longitudinal fundus images. In AREDS data, at each follow-up for each eye, there is a fundus image along with multiple manually extracted image features graded by a medical center, for example, the size of the abnormal area (drusen) and the number of drusens of a fundus image. Our analysis data set includes 53,076 eye-level observations from 7,865 eyes of 4,335 subjects. %We conducted the eye-level analysis without considering the correlation between the two eyes of the same individual.

Our predictors include three baseline demographic variables (age at enrollment, educational level, and smoking status) and the longitudinal fundus images. We used the DeepSeeNet to model fundus images, which is the CNN part in Figure \ref{workflow}(a). DeepSeeNet uses the structure of ResNet50 \citep{he2016deep}, which is a well-trained deep neural network for AMD fundus images \citep{peng2019deepseenet,peng2020predicting}. As shown in Figure \ref{workflow}(a), the hidden layer of DeepSeeNet was concatenated with the other baseline covariate nodes to feed to the survival neural network. As a comparison, we also fitted the standard time-independent CoxSNN on the baseline fundus images. Moreover, four dynamic prediction approaches using seven longitudinal AMD image features (maximum drusen size, pigmentary abnormalities, soft drusen, calcified drusen, reticular drusen, drusen area, increased pigment, see Table \ref{postcomparisontable}) were also fitted: LM-RSF, JM, tdCoxPH, and tdCoxSNN. 
Note that using these low-dimensional AMD image features is expected to produce a superior prediction result than using the fundus images directly, as they are the key measures used to determine AMD severity in clinical practice. We performed a 5-fold cross-validation. All models were trained on four folds and evaluated on the rest one fold. BS and cdAUC were evaluated on the $\Delta t = 1,\dots, 5$ windows from two landmark times, 2 years and 5 years. Specifically, to assess the prediction accuracy at each landmark time, our evaluation used subjects in the test data who were free of late-AMD at the landmark time (those who developed late-AMD or censored before the landmark time were removed from the test data).

% seven longitudinal manually graded image features, which were found significant from a multi-variable baseline Cox regression model. 

For the four methods using the seven manually graded image features, figure \ref{RealData5foldboxplot_LM} (a) shows that the tdCoxSNN outperforms the other three methods at landmark time $s=2$ years (except cdAUC at $\Delta t = 1$). At landmark time $s=5$ years, all four methods produce very similar BS values and JM gives the largest cdAUC, followed by tdCoxSNN. The improved discrimination ability of JM at year 5 may come from the better trajectory fitting as more longitudinal observations were collected and used for the longitudinal sub-model. tdCoxSNN directly using longitudinal images produces similar BS values as compared to tdCoxSNN using seven longitudinal image features. It also outperforms LM-RSF using longitudinal image features in terms of cdAUC. In practice, grading the image features is labor intensive and requires special medical image expertise. As a comparison, this prediction model (tdCoxSNN with fundus images) can directly handle the high-dimensional fundus images and achieve comparable prediction accuracy. Furthermore, when directly using fundus images, tdCoxSNN generally outperforms CoxSNN in both metrics. This highlights the value of using longitudinal fundus images to enhance the accuracy when predicting the probability of progressing to late-AMD.

To further interpret the tdCoxSNN model, we generated the saliency map \citep{yan2020deep} to visualize the regions with the most significant impact on the predicted risk score for a given subject. Figure \ref{RealData5foldboxplot_LM} (b) displays a fundus image from a participant's left eye at year 2.3 (since being on the AREDS study) where the image shows multiple large drusens (i.e., the yellow spots). Our model predicts this eye will develop late AMD with a high probability in three years (i.e., predicted probability is 70.5\% at $\Delta t=3$ year) and the truth is this eye developed late AMD 2.7 years later. The lightened dots in the saliency map indicate the informative areas in the fundus image for disease progression prediction, which expand beyond merely yellow drusen areas.

\begin{table}
\caption{\label{postcomparisontable}Characteristics of high-risk and low-risk groups in the test data.}
\centering
\begin{tabular}{lccc}
\hline
\textbf{Eye-level   variables}               & High-Risk    & Low-Risk     &                  \\
Baseline manually extracted image   features & (N = 607)    & (N = 975)    & \multirow{-2}{*}{p-value}          \\
\hline
\hspace{5mm}Maximum Drusen Size (mean ±   s.d.)          & 3.83 ± 1.18  & 2.41 ± 1.02  & \textless{}0.001 \\
\hspace{5mm}Pigmentary Abnormalities (N,   \%)           & 161 (26.5) & 64 (6.6)   & \textless{}0.001 \\
\hspace{5mm}Soft Drusen (N, \%)                          & 502 (82.7) & 308 (31.6) & \textless{}0.001 \\
\hspace{5mm}Calcified Drusen (N, \%)                     & 18 (3.0)   & 0 (0)        & \textless{}0.001 \\
\hspace{5mm}Reticular Drusen (N, \%)                     & 21 (3.5)   & 2 (0.2)    & \textless{}0.001 \\
\hspace{5mm}Drusen area (mean ± s.d.)                    & 4.39 ± 2.25  & 1.36 ± 1.36  & \textless{}0.001 \\
\hspace{5mm}Increased Pigment (N, \%)                    & 311 (51.2) & 108 (11.1) & \textless{}0.001 \\
\hline
\textbf{Subject-level   variable}               & High-Risk    & Low-Risk     &                  \\
Exclude subjects with two eye of disparate risk & (N = 283)    & (N = 440)    & \multirow{-2}{*}{p-value}          \\
\hline
Baseline age, year (mean ± s.d.)                & 70.5 ± 5.6           & 68.4 ± 4.4           & \textless{}0.001     \\
Female (N, \%)                                  & 150 (53.0)           & 261 (59.3)           & 0.110                \\
Educational level at least highschool (N,   \%) & 167 (59.0)           & 305 (69.3)           & 0.006                \\
Baseline smoking status (N, \%)                 &                      &                      & 0.516                \\
\hspace{5mm}Never smoked                                    & 114 (40.3)           & 196 (44.5)           &                      \\
\hspace{5mm}Former smoker                                   & 146 (51.6)           & 209 (47.5)           &                      \\
\hspace{5mm}Current smoker                                  & 23 (8.1)             & 35 (8.0)             &                     \\
\hline
\multicolumn{4}{l}{* Two-sample t-test for continuous features, Pearson's Chi-squared test for categorical features}\\
\end{tabular}
\end{table}

\begin{figure}
\centering
\includegraphics[width=0.7 \linewidth]{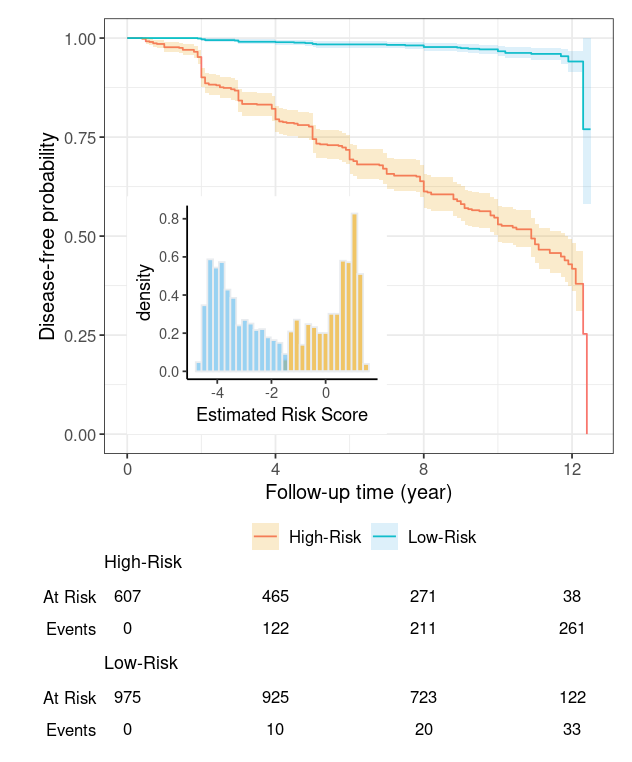}
\caption{The KM estimators of the disease-free probabilities (predicted by tdCoxSNN in the test dataset) for high-risk and low-risk groups identified by the baseline fundus image. The histograms show the estimated baseline risk score with subgroups identified by the Gaussian mixture model (log-rank test $p<2\times 10^{-16}$).}
\label{postcomparison}
\end{figure}

It is also of interest to identify individuals with a high risk of developing late AMD through the estimated risk score $\hat g_i$, which is the output of tdCoxSNN. For an illustrative purpose, we used the training and test data from the first cross-validation split. The baseline risk scores for 1,582 eyes in the test data were estimated using their baseline fundus images and three demographic variables. We then fit a Gaussian mixture model on these 1,582 estimated risk scores and identified two subgroups (see the histogram inside Figure \ref{postcomparison}). We further compared the KM curves of these two subgroups. At year 12, 43\% of subjects in the high-risk group developed late AMD, whereas less than 4\% of subjects in the low-risk group developed late AMD. We further compared the seven manually extracted image features between the two groups (Table \ref{postcomparisontable}). The individuals in the high-risk group exhibited larger maximum drusen size, more pigmentary abnormalities, more drusens (soft, calcified, and reticular), larger drusen area, and increased pigment ($p<0.001$ for all seven features). Moreover, the high-risk group individuals were older and less educated than the low-risk group individuals. All these findings consistently align with previous publications in AMD research.

\subsection{Application to PBC2 data}\label{s:pbc2}
We also applied tdCoxSNN to a publicly available dataset with low-dimensional longitudinal predictors. The data were collected between 1974 and 1984 for the research of primary biliary cirrhosis (PBC) disease \citep{fleming2011counting}. The dataset consists of 312 subjects (1,912 visits), with the event of interest being time-to-liver transplant and a censoring rate of 55\%. The predictors include 12 longitudinal variables (7 continuous lab tests, such as albumin, and 5 categorical variables, such as liver enlargement) and 3 baseline variables (e.g., sex, age at start-of-study, and treatment indicator). 

Three discrete-time deep-learning models have already been used to analyze this dataset in the past in \citet{putzel2021dynamic}. Linear-$\Delta$ and RNN-$\Delta$ are discrete-time dynamic prediction models based on the recurrent neural network (RNN) structure \citet{putzel2021dynamic}. The Dyn-DeepHit \citep{lee2019dynamic} has also been used to analyze this data. For a fair comparison, we followed the same data processing steps from \citet{sweeting2011joint} and fit tdCoxSNN by treating the discretized times as continuous times at which the longitudinal variables were measured. Table \ref{pbc2} presents the prediction accuracy in terms of dynamic BS and c-index of the four methods across different landmark times. The dynamic C-index is the C-index calculated at each landmark time based on the subjects surviving the landmark time in the test data.
The accuracy metrics for the three discrete-time deep learning models were obtained directly from \citet{putzel2021dynamic}. Each entry represents an average accuracy calculated at 2, 4, 6, and 8 months after the landmark time. In this application, our tdCoxSNN is superior to the three existing discrete-time dynamic prediction methods in four out of the six accuracy measures, while the other three methods are only one or two times the best out of the six accuracy measures.

\begin{table}
\caption{Prediction accuracy in terms of BS and c-index of deep-learning methods on PBC2 data}\label{pbc2}
\centering
\begin{tabular}{@{}lllllll@{}}
\hline
                             & \multicolumn{3}{c}{Dynamic BS}                   & \multicolumn{3}{c}{Dynamic c-index}           \\ \cmidrule(lr){2-4} \cmidrule(lr){5-7} 
Landmark time                & 4              & 7              & 10             & 4             & 7             & 10            \\ \hline
Linear-$\Delta$ [Putzel, 2021] & 0.127          & 0.130          & 0.142          & 0.80          & 0.79          & \textbf{0.89} \\
RNN-$\Delta$ [Putzel, 2021]    & 0.114          & 0.119          & 0.123          & 0.80          & \textbf{0.80} & 0.80          \\
Dyn-DeepHit [Lee, 2019]                  & \textbf{0.105} & 0.125          & 0.102          & \textbf{0.81}          & 0.68          & 0.57          \\
% LM-tdCoxSNN                    & 0.111          & 0.108 & 0.103          & 0.80 & 0.71          & 0.87         \\
tdCoxSNN                      & 0.112          & \textbf{0.093}          & \textbf{0.100} & \textbf{0.81}          & \textbf{0.80}          & 0.85          \\ \hline
% RF-LM                      & 0.145          & 0.092          & 0.109 & 0.87          & 0.88          & 0.92          \\ \hline
\end{tabular}
\end{table}

% We observed that tdCoxSNN outperformed LM-tdCoxSNN in terms of both BS and dynamic C-Index, suggesting that the proposed method could benefit from retaining more samples during training, especially given the relatively small total sample size (1,912 visits in total).

\section{Discussion}
\label{s:discuss}
We proposed a time-dependent Cox survival neural network, namely tdCoxSNN, to develop a dynamic prediction model on a continuous survival time scale with longitudinal predictors. This method allows users to directly include high-dimensional longitudinal features such as images as model inputs by incorporating other existing neural nets on top of tdCoxSNN. Such an end-to-end approach makes it possible to build a dynamic prediction model that fully utilizes these complex longitudinal biomarkers (e.g., MRI images, omics data, etc.) through a unified neural network structure. The prediction performance of tdCoxSNN is robust and satisfactory in all simulations and real data analyses. 

Compared with discrete-time dynamic prediction models \citep{lee2019dynamic,tanner2021dynamic,lin2022deep}, our model does not require the selection of time intervals for discretization, which makes it easier to process the longitudinal data for model fitting. It is worth noting that the same structure and hyper-parameters for tdCoxSNN in this study worked well through all analyses. Additional tuning of the hyper-parameters may further improve the prediction accuracy of our model. The computational cost of tdCoxSNN is low since the loss function in the training procedure becomes stable within 20 epochs (e.g. within 5 minutes in each simulation run). 

There are a few limitations in the current version of tdCoxSNN. First, tdCoxSNN does not model the trajectory of longitudinal predictors and assumes constant step functions between time intervals, which may restrict the long-term predictive accuracy of the model. A possible solution is to model the continuous-time hidden dynamics of risk score $g(X_i,y_i(t))$ over time through the ordinary differential equations \citep{rubanova2019latent}. This will relax the constant step function assumption but requires fairly frequent measurements of longitudinal predictors with additional computational costs. Second, our method assumes time-invariant effects of predictors on the survival outcome, which is different from the LM method (the effects of the predictors change when different landmark times are chosen in the LM approach). This can be relaxed by including time as an additional input in the neural network (e.g., $g_\theta(X_i,y_i(t),t)$). Such modification allows the model to account for changes in predictor effects over time. This strategy has been applied in the deep learning survival model for time-invariant covariates \citep{kvamme2019time}. 

\section*{Data Availability}
The PBC2 data used in this work can be found in \url{https://rdrr.io/cran/joineRML/man/pbc2.html} from the R package \{joineRML\} \citep{hickey2018joinerml}.
The AREDS data used in this work are available from the online repository dbGaP (accession: phs000001.v3.p1).

\section*{Acknowledgments}
The research presented in this paper was supported by NIH/NIGMS under Grant Number R01GM14107 and NIH/NEI under Grant Number R21EY030488. This research was also supported in part by the University of Pittsburgh Center for Research Computing through the resources provided. Specifically, this work used the HTC cluster, which is supported by NIH award number S10OD028483.

% \section*{Supplementary Materials}

% Web Appendix A, referenced in Section~\ref{s:model}, is available with
% this paper at the Biometrics website on Wiley Online
% Library.\vspace*{-8pt}

\bibliographystyle{rss}
\bibliography{tdCoxSNN}
\end{document}